\documentclass[runningheads]{llncs}
\usepackage{graphicx}
\usepackage{algorithm}
\usepackage{algpseudocode}
\usepackage{caption}
\usepackage{subcaption}
\usepackage{comment}
\usepackage{amsmath}
\usepackage{booktabs}
\usepackage{url}
\usepackage[flushleft]{threeparttable}
\begin{document}
\title{Sim-to-Real Deep Reinforcement Learning with Manipulators for Pick-and-place\thanks{Supported by EPSRC project No. EP/S03286X/1, EPSRC RAIN project No. EP/R026084/1, EPSRC RNE project No. EP/P01366X/1 and UKAEA/EPSRC Fusion Grant 2022/2027 No. EP/W006839/1.}}
\titlerunning{Sim-to-Real DRL with Manipulators for Pick-and-place}
%
\author{Wenxing Liu\inst{1,2}\orcidID{0000-0002-3195-3862} \and
Hanlin Niu\inst{1,2}\orcidID{0000-0003-0457-0871} \and Robert Skilton\inst{1}\orcidID{0000-0003-1076-906X} \and
Joaquin Carrasco\inst{2}\orcidID{0000-0002-7499-6408} }
\authorrunning{W. Liu et al.}
%
\institute{Remote Applications in Challenging Environments (RACE), United Kingdom Atomic Energy Authority, Culham, UK \\
\email{\{wenxing.liu, hanlin.niu, robert.skilton\}@ukaea.uk}
\and
Department of Electrical \& Electronic Engineering, The University of Manchester, Manchester, UK \\
\email{joaquin.carrasco@manchester.ac.uk}}
\maketitle              
\begin{abstract}
When transferring a Deep Reinforcement Learning model from simulation to the real world, the performance could be unsatisfactory since the simulation cannot imitate the real world well in many circumstances. This results in a long period of fine-tuning in the real world. This paper proposes a self-supervised vision-based DRL method that allows robots to pick and place objects effectively and efficiently when directly transferring a training model from simulation to the real world. A height-sensitive action policy is specially designed for the proposed method to deal with crowded and stacked objects in challenging environments. The training model with the proposed approach can be applied directly to a real suction task without any fine-tuning from the real world while maintaining a high suction success rate. It is also validated that our model can be deployed to suction novel objects in a real experiment with a suction success rate of 90\% without any real-world fine-tuning. The experimental video is available at: \url{https://youtu.be/jSTC-EGsoFA}.


\end{abstract}
\section{Introduction}
Robotic technology has resulted in significant advancements in various areas such as goal attainment \cite{9833460}, object manipulation \cite{wang2021optimal,10161062}, formation tracking \cite{wu2022mixed,wu2023finite}, human-robot interaction \cite{8967834}, collision avoidance \cite{na2023federated,na2022bio}, and path planning \cite{niu2020energy}.
 Deep reinforcement learning (DRL) has become an essential element in robotic control, where an agent gradually develops a particular strategy through interaction with the environment to receive maximum rewards. However, transferring DRL models from simulation to the real world is challenging due to the discrepancy between the two environments. It takes considerable time to fine-tune the model parameters to adapt to the real-world environment. The use of robotic arms in a real environment is time-bound, raising the crucial question of reducing the real-world fine-tuning duration while maintaining high accuracy in picking and placing objects.
\vspace{-0.3cm}
\begin{figure}[!ht]
\centering
\subfloat[]{\includegraphics[width=2in,height=1.6in]{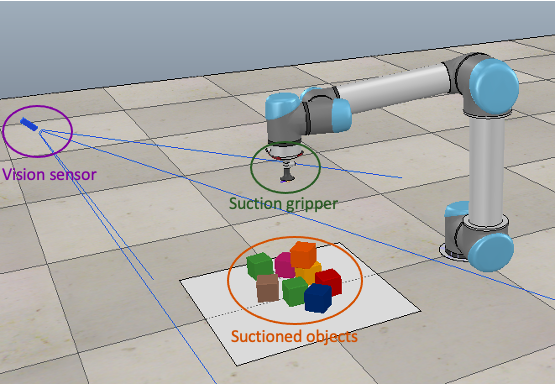}}
\hspace{0.05cm}
\subfloat[]{\includegraphics[width=2in,height=1.6in]{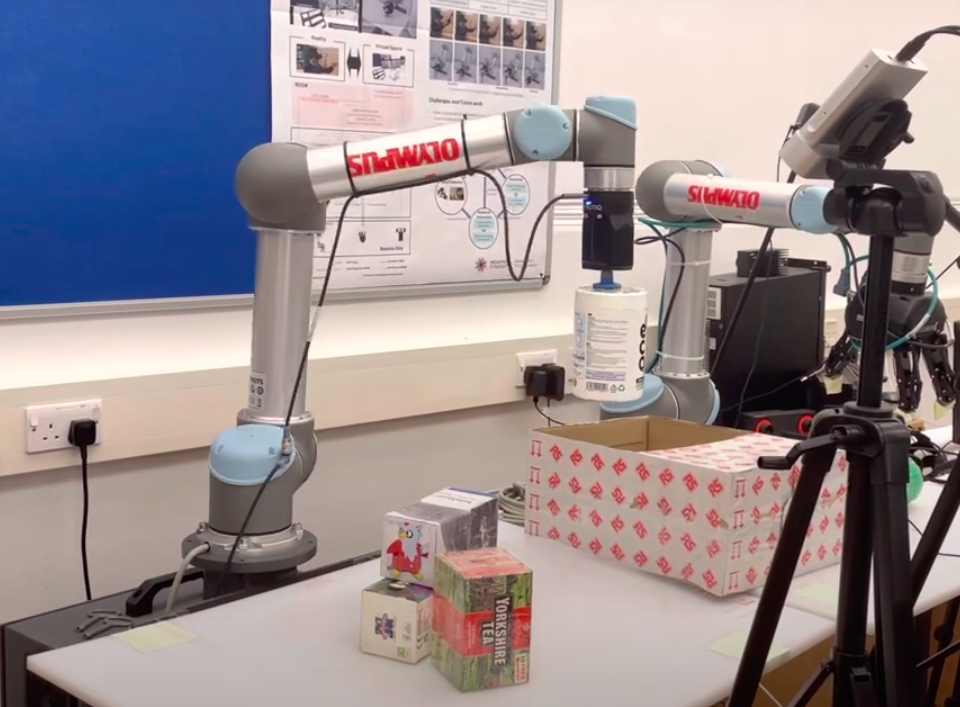}}

\caption{(a) The training environment in simulation; (b) Pick-and-place objects with the proposed method in the real world}
    \label{pp}
\end{figure}
\vspace{-0.4cm}

In comparison to conventional methods for object manipulation, using DRL poses certain challenges, including the need for feature extraction through neural networks to enable suctioning of objects, the ability to generalize to novel objects with different shapes and heights, developing self-supervised approaches to avoid the need for pre-labelled training data, and adapting to challenging environments such as crowded and stacked objects. This paper proposes a self-supervised end-to-end DRL approach that enables robots to effectively and efficiently pick and place objects by directly transferring the training model from simulation to the real world, see Fig.~\ref{pp}. The key contributions of the paper can be summarized as follows:
\begin{enumerate}
\item A fully self-supervised DRL method that utilizes visual information to enable manipulators to learn how to pick and place objects is proposed. By encouraging the Universal Robot 5 (UR5) robot arm to suction the area near the centre of the target object in simulation, the proposed training model can directly be applied to pick and place trained objects in the real world with a $90\%$ suction success rate.

\item In particular, a height-sensitive action policy is developed to facilitate the proposed self-supervised vision-based DRL method for suctioning in a challenging environment where objects are crowded and stacked.

\item The efficacy of the proposed method is evaluated in both simulated and real environments, demonstrating its robustness and applicability to a variety of scenarios. The approach is also shown to achieve a $90\%$ suction success rate on novel objects without requiring any fine-tuning in the real world.

\end{enumerate}

\section{Related Work}
\textbf{Sim-to-real:} The concept of Sim-to-real training has been extensively researched to minimize the gap between simulation and real-world environments \cite{kaushik2022safeapt}. The central concept is to modify simulated environments using real-world samples \cite{tobin2017domain}. In \cite{tzeng2020adapting}, a novel domain adaptation approach was proposed for robot perception to close the reality gap between simulation and the real world by searching common features of synthetic and real data. In \cite{planche2017depthsynth}, an end-to-end pipeline was developed to generate realistic depth data from 3D simulation models by accurately modelling vital real-world factors such as sensor noise and surface geometry. In recent years, there have been publications where only simulation was used for training, yet they performed well in the real environment. A grasp quality convolutional neural network was developed in \cite{mahler2017dex} to configure the robustness of grasp from a point cloud. In \cite{viereck2017learning}, a closed-loop controller was trained with only simulation to make robots tackle unexpected changes in objects. Switching from RGB images to depth data can help reduce the sim-to-real gap, as depth images carry less information than RGB images. Due to limitations in physical properties, such as the inability to accurately capture dark-coloured or thin objects, depth cameras may struggle to measure certain objects, thereby hindering the real-world performance of robotic arms. The proposed method is capable of suctioning objects of various shapes and heights, making it more versatile. \\

\noindent \textbf{Pick-and-place:} 
Pick-and-place is a crucial concept in the field of robotic manipulators \cite{perumaal2013automated,kumar2014object,huang2018case}. In recent times, there has been an increasing focus on the manipulation of objects through picking and placing. For instance, an industrial robot system with several stationary cameras was mentioned in \cite{gecks2005human} to ensure safe human-robot cooperation. In \cite{qul2019edge}, an application of visual serving to a 4 degree-of-freedom (DOF) robot manipulator was proposed to pick and place a target using the edge detection method as visual input. A remote-controlled mobile robot was developed and designed in \cite{abdulkareem2019design} to deal with the pick-and-place task. In \cite{sharan2012client}, the software development for a vision-based pick-and-place robot was presented to provide the computational intelligence required for its operation. In order to eliminate human intervention or error and work more precisely, a pick-and-place robot using Robo-Arduino was developed in \cite{harish2017pick} for any pick-and-place functions. A pick-and-place robot which offered sensing, control and manufacturing assistance was present in \cite{smys2019robot}, which improved productivity and reduced the risk of injury because of repetitive tasks. However, the studies mentioned above are primarily focused on model-based grasping using certain types of robotic manipulators, which could be seen as a constraint. The approach presented in this paper emphasizes end-to-end and pixel-to-pixel suctioning, which can be readily adapted to other robotic systems.

\section{Methodology}
The proposed method is trained entirely through self-supervision, which involves the interaction between the UR5 robot arm and the simulated environment.

\subsection{System Overview}

\begin{figure*}
        \centering
        \includegraphics[width=\linewidth]{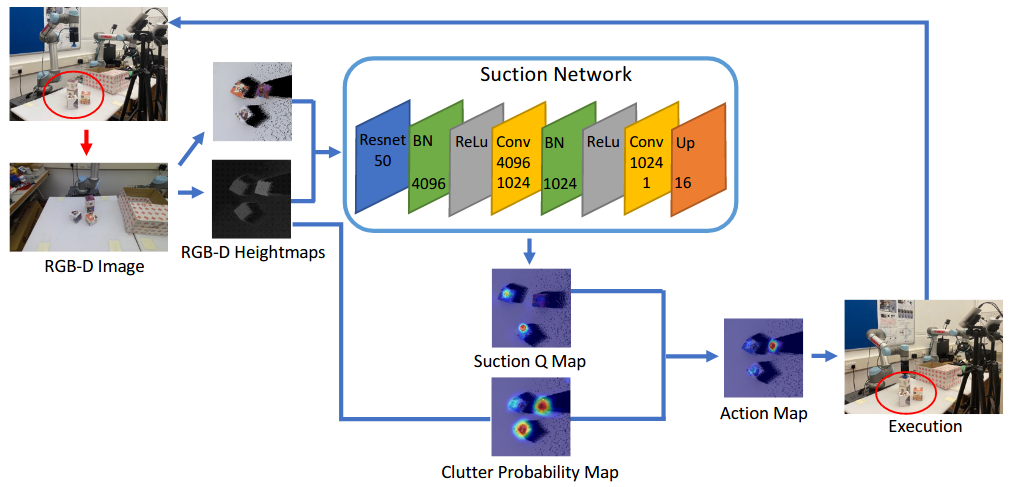}
        \caption{Overview of the proposed framework. BN stands for Batch Normalization. Conv represents convolution. Up stands for upsampling. The red circle denotes pixel-wise best suction position. More details can be found from Algorithm \ref{algorithm1}.}
        \label{sim}
\end{figure*}
\vspace{-0.4cm}
The overview of the proposed DRL framework is shown in Fig.~\ref{sim}. The RGB-D image $g_t$, captured by a fixed camera, is orthographically projected in the gravity direction to construct the colour heightmap $c_t$ and the depth heightmap $d_t$. Then both heightmaps are fed into the suction network to generate a suction Q map $q_t$. By detecting different heights from $d_t$, a clutter probability map $l_t$ can be obtained. The position with the highest probability in the action map denotes the pixel-wise best suction position $[x_t, y_t]$. The suction height $z_t$ is obtained from $d_t$.

\subsection{Reward Space and State Space}
\subsubsection{Reward Space} During iteration $t$, the centre distance $\psi_t$ between the suctioned object and the pixel-wise best suction position $[x_t, y_t]$ can be computed by 
\begin{equation} 
\ \psi_t = \sqrt{(x_t-\sigma_t)^2+(y_t-\iota_t)^2}
\end{equation}
where $\sigma_t$ and $\iota_t$ denote $x,y$ positions of the centre of the suctioned object.

The reward function can be defined as follows:
\begin{equation}\label{rcc}
    r = \frac{r_p}{(\psi_t+\delta)} r_g 
\end{equation}
where $r_p$ is a positive constant reward, $\delta$ is a small positive number which prevents zero division, $r_g=1$ if the object is successfully suctioned, otherwise $r_g=0$. As a result, the reward stimulates agents to suction the area close to the centre of the expected suctioned object during each iteration, which increases the suction success rate. 
\subsubsection{State Space} The state $s_t$ contains the colour heightmap $c_t$ and the depth heightmap $d_t$.

\subsection{Network Structure}
As shown in Fig.~\ref{sim}, the input of the suction network passes data through ResNet-50 \cite{szegedy2017inception} to extract features from both heightmaps. Then the aforementioned features are fed into a Batch Normalization layer \cite{paszke2019pytorch} with 4096 input features, a ReLu layer \cite{paszke2019pytorch}, a Convolution layer \cite{paszke2019pytorch} with 4096 input channels and 1024 output channels. After passing data through another Batch Normalization layer \cite{paszke2019pytorch}, ReLu layer \cite{paszke2019pytorch} and Convolution layer \cite{paszke2019pytorch}, data are processed by a bilinear upsample layer \cite{paszke2019pytorch} with a scale factor of 16. The output of the suction network shares the same image size as the heightmap input, which is a dense pixel-wise Q map. The pixel with the highest probability in the action map denotes the best suction position.

\subsection{Height-sensitive Action Policy} 
To effectively suction in a challenging environment, a height-sensitive action policy is proposed. As can be seen from Fig.~\ref{sim}, $q_t$ can be acquired from $c_t$ and $d_t$. However, the information contained in $q_t$ is not enough to make the proposed framework sensitive to the heights of the grasped objects. As a result, we introduce the clutter probability map $l_t$. The depth heightmap is shifted along one axis for 60 pixels to generate a translated map. By contrasting the depth difference between the translated and the original depth heightmap, the pixel with enough depth difference is counted as 1 otherwise 0, which builds the clutter probability map $l_t$. The action space $a_t$ can be computed as follows:

\begin{equation}
    a_t= \arg \max_{a}(q_tl_t)
\end{equation}
where $q_t$ and $l_t$ depend on the action.
\begin{algorithm}[!t]
\caption{Vision-based DRL for Pick-and-place} \label{algorithm1}
\begin{algorithmic}[1] 
\State {Initialize training parameter $\phi_t$, learning rate $\alpha$, RGB-D image $g_t$,  discounted factor $\gamma$, training steps parameter $T$, replay buffer $R_p$.}

\While {$t < T$} 

\State{Generate $c_t$ and $d_t$ from $g_t$.}

\State{Generate $l_t$ from $d_t$.}

\If {object number $b_t <$ empty threshold} 

\State{Feed $c_t$ and $d_t$ into the suction network with the height-sensitive action policy to get action-value function $Q(\phi_t,s_t,a_t)$.}

\If {$t > 2$} 

\State{Use $Q(\phi_{t-1},s_{t-1},a_{t-1})$ to generate $r_t$.}

\State{Minimize the temporal difference error $\xi_{t-1}$:\\
$\quad \quad \quad \quad \quad \quad \quad  y_{t-1} = r_{t} + \gamma \underset{a}{\max}(Q(\phi^-_{t-1},s_{t},a))$.\\
$\quad \quad \quad \quad \quad \quad \quad \xi_{t-1} = Q(\phi_{t-1},s_{t-1},a_{t-1})-y_{t-1}$.}

\State{Sample a minibatch from $R_p$ for experience replay.}

\EndIf

\State{Suction objects.}

\State{Store $(c_t, d_t, a_t)$ in $R_p$.}

\Else
\State{Reposition objects.}

\EndIf

\EndWhile

\end{algorithmic} 
\end{algorithm}

\section{Experiments and Results}
The feasibility of the proposed method is validated in both simulated and real environments. The proposed approach is implemented on a desktop with Nvidia GTX 2080 and Intel Core i9 CPU with 64 GB RAM. 

\subsection{Training Details}
The proposed method is trained in Coppeliasim \cite{rohmer2013v} using Python \cite{lutz2001programming} and Pytorch \cite{paszke2019pytorch}. The UR5 robot arm is connected with a suction gripper \cite{ge2021supervised} to pick and place objects, as shown in Fig.~\ref{pp} (a). During each training iteration, a vision sensor captures RGB-D images of the UR5 robot arm in a $0.448\times0.448$ m$^2$ workspace. The resolution of the RGB-D images is $640\times480$. The UR5 robot arm motion planning task can be accomplished using Coppeliasim \cite{rohmer2013v} internal inverse kinematics. The suctioned objects are $5\times5\times5$ cm$^3$ cubes.

Depending on the size of the suction gripper and the intrinsic of the vision sensor, we set $r_p = 15$ and $\delta=0.00001$ in \eqref{rcc}. For other robotic platforms, these values can also be reconfigured. In Algorithm \ref{algorithm1}, the learning rate $\alpha$ is set to 0.0001. The discounted factor $\gamma$ has a fixed value of 0.5. The training steps parameter $T$ is set to 400. The training is considered to be successful if the UR5 robot arm is able to pick and place target objects which are randomly dropped into the workspace.

\subsection{Evaluation Metrics}
We design two metrics to evaluate the suction performance of the UR5 robot arm. For all these metrics, a larger value leads to better performance.

The suction success rate $S_r$ is given by:
\begin{equation}
    S_r = \frac{N_s}{N_i}\times100\%
\end{equation}
where $N_s$ stands for the number of successful suctions, $N_i$ represents the number of training steps. 

The distance rate $D_r$ is defined as follows:
\begin{equation}
    D_r = \frac{N_d}{N_i}\times100\%
\end{equation}
where $N_d$ is the number of times when $\psi_t < 0.015$ m.

\subsection{Baseline Method}
The performance of our system is compared with the following baseline approach:
\textbf{Visual Grasping method} shares the same input as our proposed method to generate the probability maps for best suction positions. However, it takes binary classification for the reward space design in which 1 stands for successful grasp and 0 otherwise. This baseline method is analogous to the Visual Pushing Grasping (VPG) method \cite{zeng2018learning}. Nevertheless, we extend this method to our suction framework for a fair comparison. 

\subsection{Simulation Evaluation}
\begin{figure}[!ht]
\centering
\subfloat[]{\includegraphics[width=2in,height=1.5in]{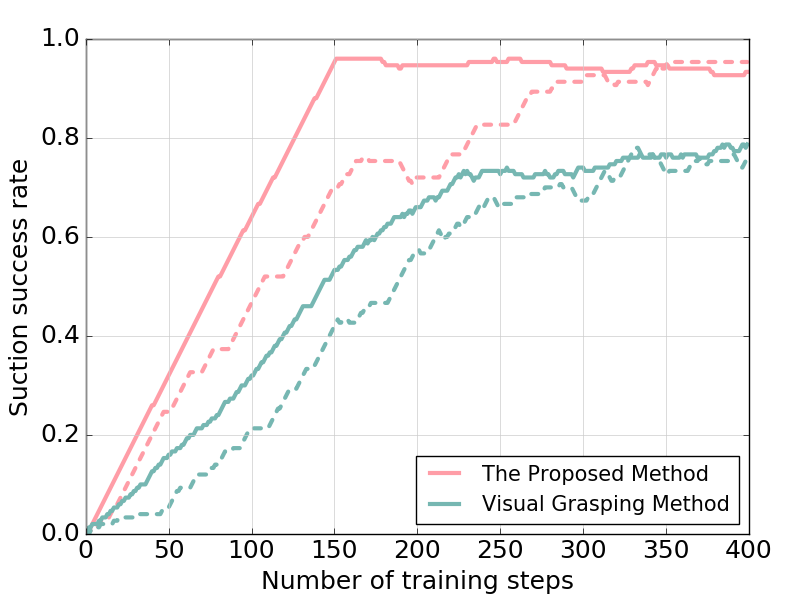}}
\hspace{0.05cm}
\subfloat[]{\includegraphics[width=2in,height=1.5in]{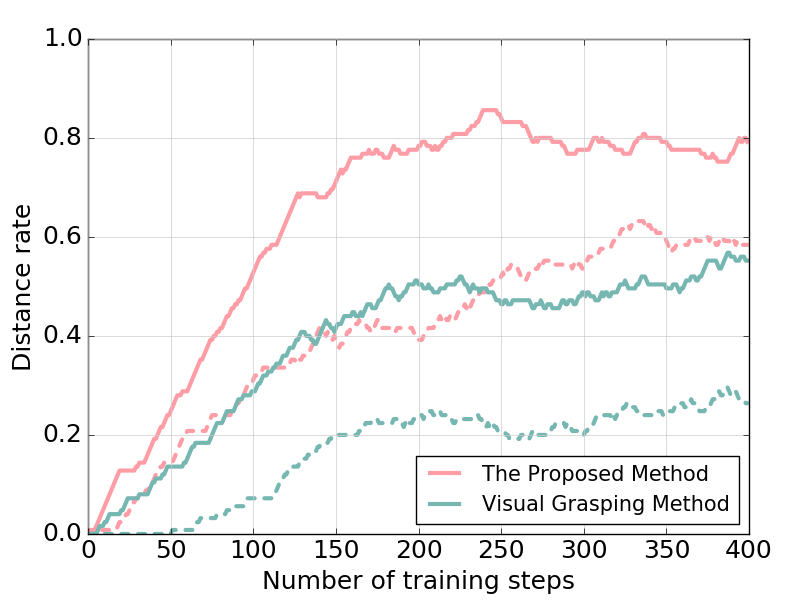}}

\caption{(a) Suction success rates; (b) Distance rates of both methods. The dotted lines represent methods without the height-sensitive action policy.}
    \label{suction}
\end{figure}

To confirm the validity of our design, we train both methods in simulation for 400 steps. Overall the proposed method outperforms the Visual Grasping method in terms of both suction success rate and distance rate by large margins. It can be obtained from Fig.~\ref{suction} (a) that the proposed method arrives at around $97\%$ suction success rate at 150 training steps, while the Visual Grasping method shows only $52\%$. Removing the height-sensitive action policy from both methods leads to a longer time to achieve the same suction success rate. As can be seen from Fig.~\ref{suction} (b), the distance rate of the proposed method reaches around $80\%$ at 400 training steps, whereas the Visual Grasping method shows only $58\%$. When the height-sensitive action policy is separated from both methods, the distance rates decrease by $20\%$ and $33\%$, respectively. These simulation results confirm the validity of the proposed reward space design in improving the suction success rate by encouraging robots to suction the area close to the centre of the expected suctioned object.

\begin{figure}[!ht]
\centering
\subfloat[]{\includegraphics[width=2in,height=1.4in]{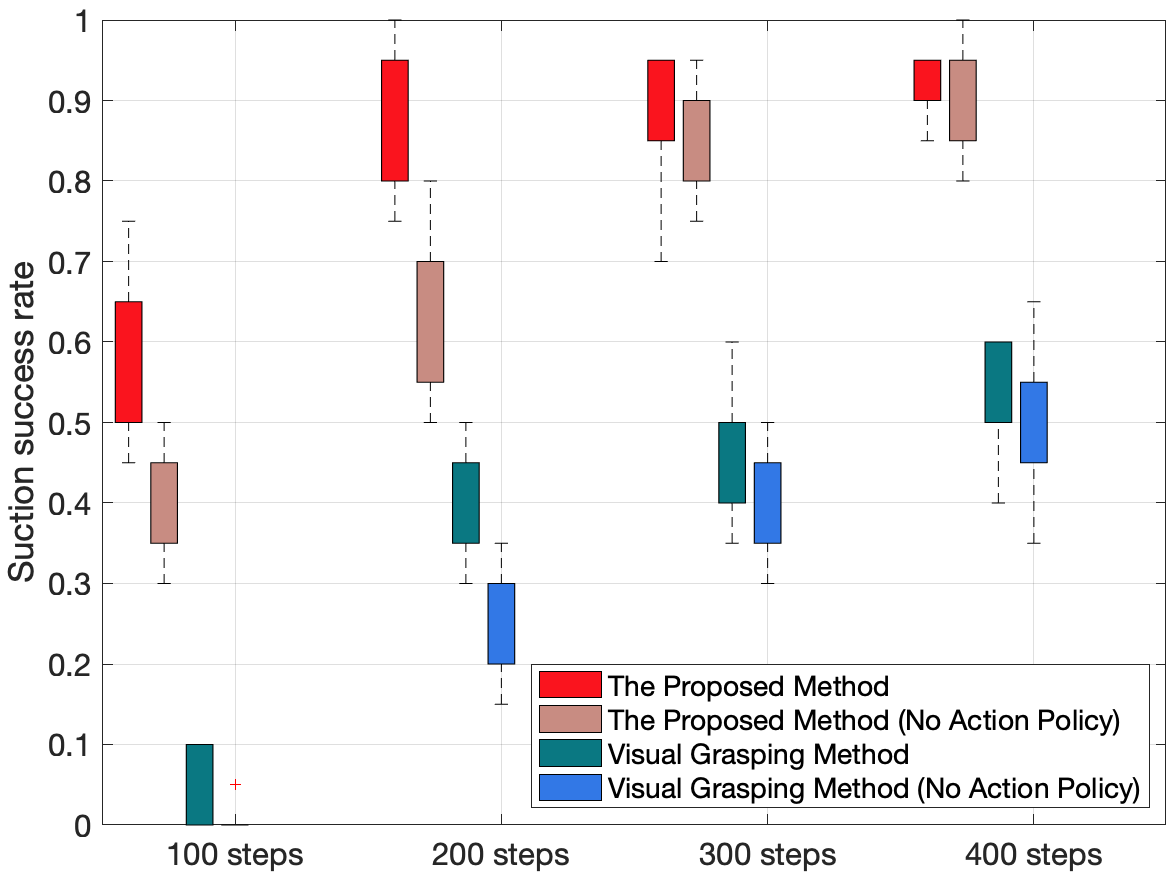}}
\hspace{0.05cm}
\subfloat[]{\includegraphics[width=2in,height=1.4in]{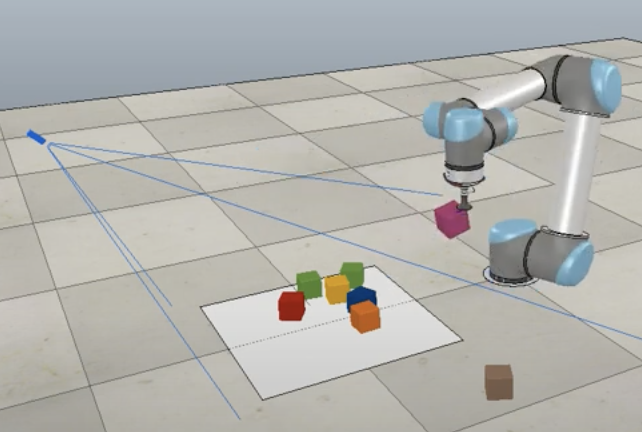}}

\caption{(a) Real-world evaluation of both methods; (b) Edge suctioning with the Visual Grasping method. Although the suctions are considered successful in simulation, they will cause real-world failures.}
    \label{vgsim}
    \vspace{-0.6cm}
\end{figure}
\subsection{Real-world Evaluation}
We evaluate both methods directly in the real environment using the models trained in Fig.~\ref{suction} without any real-world fine-tuning. The UR5 robot arm is connected to a Robotiq EPick vacuum gripper for real-world evaluation. A fixed Azure Kinect camera is used to capture RGB-D images with a resolution of $1280\times720$. The suctioned objects are $7\times7\times7$ cm$^3$ cubes. Fig.~\ref{vgsim} (a) depicts the box plot of real-world evaluation with both methods. The proposed method achieves a $90\%$ suction success rate at 200 training steps in real-world evaluation, while the Visual Grasping method shows only $40\%$. When the height-sensitive action policy is separated from both methods, the suction success rates drop to $65\%$ and $30\%$, respectively. By implementing the proposed method, the suction success rate gap between the simulation and the real environment is only $6\%$, much smaller than the gap using the Visual Grasping method ($26\%$). Although the suctions are considered successful in Fig.~\ref{vgsim} (b), they will result in real-world failures because of edge suctioning, which enlarges the gap between the simulation and the real world. The proposed method boosts the suction success rate of the UR5 robot arm by motivating it to suction the region in proximity to the centre of the intended object, outperforming the Visual Grasping method.

\subsection{Suction in Challenging Environments}
\begin{figure}[!ht]
\centering
\subfloat[]{\includegraphics[width=0.8in]{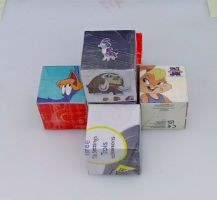}}
\hspace{0.05cm}
\subfloat[]{\includegraphics[width=0.8in]{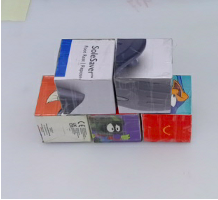}}
\hspace{0.05cm}
\subfloat[]{\includegraphics[width=0.8in]{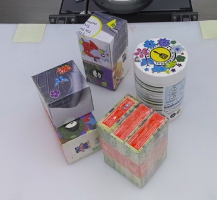}}
\caption{Suctioning in challenging environments: (a) Environment 1; (b) Environment 2; (c) Environment 3}
\label{chall3}
\end{figure}

\begin{figure}[!ht]
\centering
\subfloat[]{\includegraphics[height=1.2in]{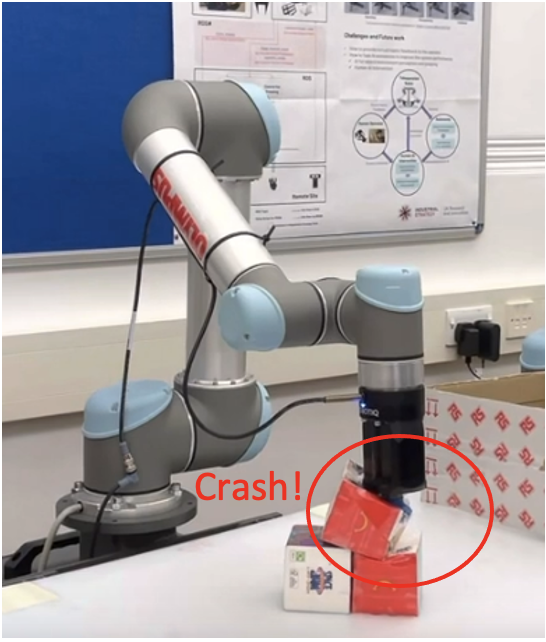}}
\hspace{0.05cm}
\subfloat[]{\includegraphics[height=1.2in]{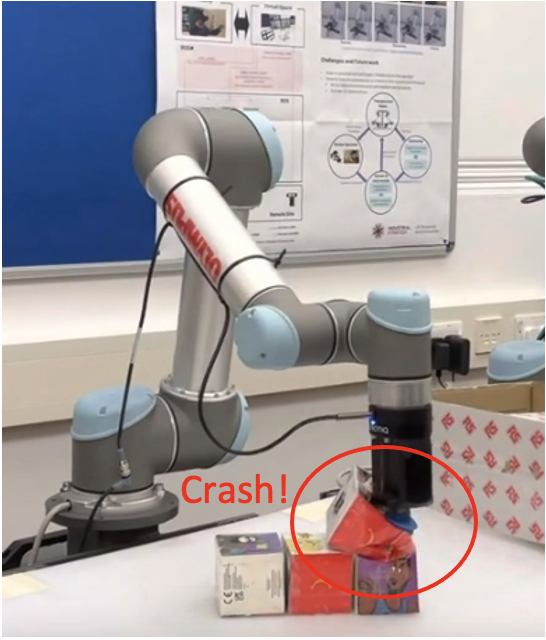}}
\hspace{0.05cm}
\subfloat[]{\includegraphics[height=1.2in]{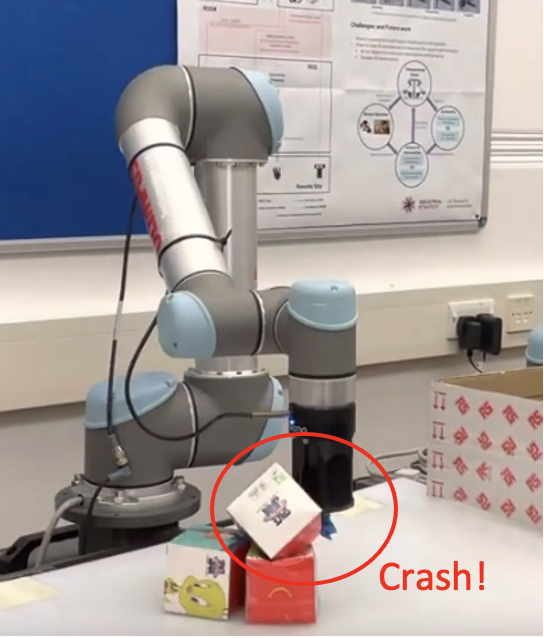}}
\caption{Examples of collisions without the height-sensitive action policy}
\label{crash}
\end{figure}

The performance of the height-sensitive action policy in our proposed method is validated in this section. 
\begin{figure*}
        \centering
        \includegraphics[width=1\linewidth]{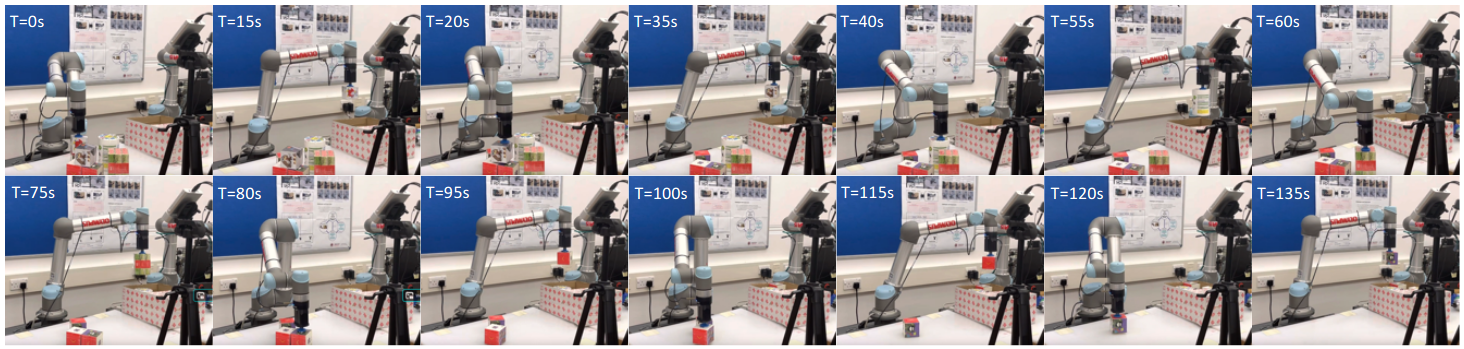}
        \caption{The demonstration of suctioning in Challenging Environment 3}
        \label{cha3}
        \vspace{-0.6cm}
\end{figure*}
The common practice among humans when picking and placing crowded or stacked objects is to first grasp the objects on the top and then those at the bottom, as it is deemed to be a safer approach. Inspired by this, we have developed a height-sensitive action policy that instructs the UR5 robot arm to take the heights of objects into consideration, thus minimizing the risk of potential collisions when applying the presented approach. The testing environments are elucidated in Fig.~\ref{chall3}. Environment 1 contains fully stacked objects. Environment 2 consists of half-stacked objects, which is more challenging. Environment 3 is the most challenging environment which contains both half-stacked objects as well as novel objects. As shown in Fig.~\ref{cha3}, the proposed method is able to handle crowded and stacked objects in a safe manner. It can be obtained from Table \ref{table41} that the more challenging the environment is, the more effective the height-sensitive action policy is. If the height-sensitive action policy is removed from both methods in Environment 3, the collision probability will increase by $45\%$ and $51\%$, respectively. Some failed examples are shown in Fig.~\ref{crash}, which are due to the fact that the UR5 robot arm tries to suction the object below first rather than the object above, thus colliding with the object above. This confirms the necessity of the proposed height-sensitive action policy which ensures safety during the entire movement of the UR5 robot arm.
\vspace{-0.4cm}
    \vspace{-1.6cm}
\begin{table}[]
\centering
\begin{threeparttable}
\caption{Average performance in challenging environments} \label{table41}
\begin{tabular}{@{}l@{\quad}c@{\quad}c@{\quad}c@{}}
\hline
\quad Collision rate (\%) &  Env 1 & Env 2 & Env 3 \ \ \quad \\
\hline
\quad The proposed method      & 0       & 0        & 5 \ \ \quad     \\
\hline
\quad The proposed method (No Policy)      & 0        & 45         & 50  \ \ \quad  \\
\hline
\quad Visual Grasping method      & 0        & 1         & 9 \ \ \quad  \\
\hline
\quad Visual Grasping method (No Policy)      & 5        & 55.5         & 60   \ \ \quad   \\
\hline
\end{tabular}
\begin{tablenotes}
   \item[*] No Policy means without the height-sensitive
   action policy.  
  \end{tablenotes}
  \end{threeparttable}
\end{table}
  \vspace{-1cm}
\subsection{Real-world Unseen Objects Challenge}
\begin{figure}[!ht]
\centering
\subfloat[]{\includegraphics[width=1in]{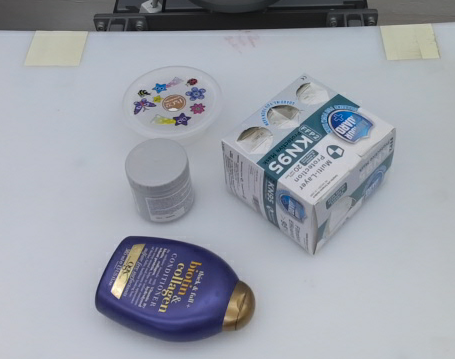}}
\hspace{0.05cm}
\subfloat[]{\includegraphics[width=1in]{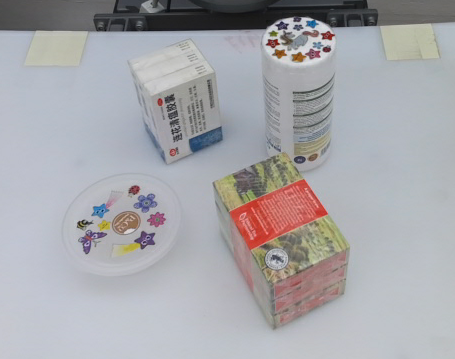}}

\caption{Novel objects for real-world suctioning: (a) Environment 1; (b) Environment 2}
\label{edge}
\end{figure}

\begin{figure}[!ht]
  \begin{center}
    \includegraphics[width=3.6in]{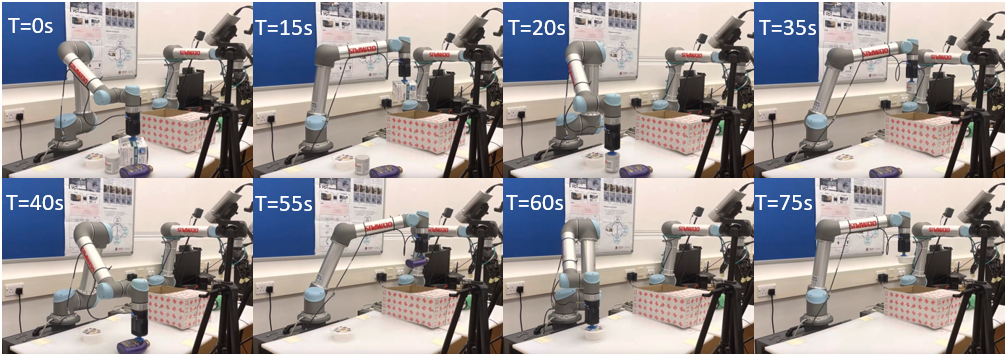}    
    \caption{Pick-and-place novel objects with the proposed method}

    \label{novelreal}
  \end{center}
    \vspace{-0.65cm}
\end{figure}

In this section, we validate the generalisation capability of our proposed vision-based DRL method. As shown in Fig.~\ref{edge}, novel objects contain cylinders of different heights as well as irregularly shaped objects. The proposed method can generalise to novel objects with a suction success rate of $90\%$ without any real-world fine-tuning. More details can be seen in Fig.~\ref{novelreal}. 

\section{Conclusion}
In this paper, we introduce a self-supervised DRL approach using vision-based methods to reduce the gap between simulated and real environments. Our proposed approach shows significant improvement over the Visual Grasping method in terms of suction success rate and distance rate. The suction success rate of the proposed approach reaches $90\%$ after 200 training steps, while the Visual Grasping method only achieves $40\%$. By implementing a height-sensitive action policy, our method can safely pick and place crowded and stacked objects in challenging environments. Our model can be directly applied to real-world experiments and is capable of generalizing to new objects with a success rate of $90\%$ without any fine-tuning. In the future, an exploration of optimizing the proposed method to handle more complicated scenarios will be carried out.

 \bibliographystyle{splncs04}
\bibliography{mybib.bib}

\end{document}